# A supervised discriminant data representation: application to pattern classification

F. Dornaika[1,2,3] · A. Khoder[2] · A. Moujahid[2] · W. Khoder[4]

## Abstract

The performance of machine learning and pattern recognition algorithms generally depends on data representation. That is why, much of the current effort in performing machine learning algorithms goes into the design of preprocessing frameworks and data transformations able to support effective machine learning. The method proposed in this work consists of a hybrid linear feature extraction scheme to be used in supervised multi-class classification problems. Inspired by two recent linear discriminant methods: robust sparse linear discriminant analysis (RSLDA) and inter-class sparsity-based discriminative least square regression (ICS_DLSR), we propose a unifying criterion that is able to retain the advantages of these two powerful methods. The resulting transformation relies on sparsity-promoting techniques both to select the features that most accurately represent the data and to preserve the row-sparsity consistency property of samples from the same class. The linear transformation and the orthogonal matrix are estimated using an iterative alternating minimization scheme based on steepest descent gradient method and different initialization schemes. The proposed framework is generic in the sense that it allows the combination and tuning of other linear discriminant embedding methods. According to the experiments conducted on several datasets including faces, objects, and digits, the proposed method was able to outperform competing methods in most cases.

## 1 Introduction

Modern systems of interest based on computer vision, such as driver assistance systems, healthcare, or surveillance systems, may be characterized as high-dimensional systems generally embedded onto low-dimensional manifolds that preserve the intrinsic properties of the original data. Learning good representations of the data able to extract and organize the discriminative information is of great interest. It may reduce the memory and computational requirements and, more importantly, tends to improve the performance of classifiers or other predictors. This explains why representation learning is becoming a hot research topic (e.g., [15, 16, 21, 22, 34, 41–43]).

Among the various ways to learn representations, this work focuses on feature selection and feature extraction. In general, a feature may be called relevant, irrelevant, or redundant. In general, a feature is called irrelevant if it does not contribute to the improvement of the predictive model and may degrade the classification accuracy when considered in the classification process. Relevant features are those features that contribute to a better prediction model and thus to higher classification accuracy. These features are the ones that the model should extract and select among all others. A redundant feature does not contribute to the model performing better in classification.

✉ F. Dornaika
fadi.dornaika@ehu.eus

1 School of Computer and Information Engineering, Henan University, Kaifeng, China

2 University of the Basque Country UPV/EHU, San Sebastian, Spain

3 IKERBASQUE, Basque Foundation for Science, Bilbao, Spain

4 Laboratoire d'Informatique et des Systèmes (LIS), Université de Toulon, Toulon, France

Feature extraction can be performed using linear or nonlinear methods. Most feature extraction methods look for a linear transformation that maps the original features to another space from which latent variables can be obtained. In these methods, feature ranking or selection can be imposed by adding a $\ell_{2,1}$-norm constraint on the projection matrix in the global criterion, which has demonstrated the robustness and joint sparsity property of the resulting projection matrix [33, 35]. For example, algorithms such as robust sparse linear discriminant analysis (RSLDA) [35] have been proposed to address many limitations of the classical linear discriminant analysis (LDA) algorithm [27], where the norm $\ell_{2,1}$ is imposed on the regularization term in the global criterion to ensure that the method performs feature ranking and weighting. This $\ell_{2,1}$ norm has been also adopted to construct supervised model where the margins of samples from the same class are greatly reduced, while the margins of samples of different classes are significantly increased [37].

Nowadays, researches focus on the use of linear projection models that perform feature selection (implicit ranking) and extraction simultaneously [35, 44]. In this sense, low-rank representation (LRR) methods provide effective subspace learning representations that have been shown to be more robust by exploring global representative structural information between samples. However, conventional LRR approaches fail to provide a low-dimensional projection of training samples with supervised information. This problem was recently approached in [13] by presenting a method that integrates the properties of LRR with supervised dimensionality reduction techniques to simultaneously obtain an optimal low-rank subspace and a discriminative projection. Other methods instead use least squares regression frameworks to achieve a discriminative feature extraction [37], or extract the optimal projection matrix using the elastic net penalty and fuzzy set theory with the locality-preserving projection technique [31]

In this paper, we present a unified and hybrid discriminant embedding method that can retain the strengths of two recent discriminant methods: (i) robust sparse linear discriminant analysis (RSLDA) [35] and (ii) inter-class sparsity-based discriminative least square regression (ICS_DLSR) [37]. The former promotes linear discriminant analysis with implicit feature selection, and the latter promotes inter-class sparsity, which means that the projected features share a common sparse structure for the samples in each class.

Thus, the main contributions are as follows. First, we deduce a novel objective function to estimate the linear transformation which has proven to refine the solution of RSLDA (projection matrix **Q**). Second, we provide an optimization algorithm in which the linear transformation is estimated by a gradient descent method. This allows setting the initial projection matrix to a hybrid combination of transformation matrices obtained from both ICS_DLSR [37] and RSLDA [35]) methods. Finally, we propose two initialization procedures for the linear transformation, which lead to two variants of the proposed algorithm.

Indeed, our approach takes advantage of two powerful discriminative methods at two levels: (1) the initialization of the hybrid linear transformation and (2) the refinement by the proposed single new criterion. The proposed method is also able to obtain a well-constructed projection space that ensures high classification accuracy. It can also be used to tune an already obtained projection matrix. Our approach can be generic in the sense that any hybrid initial projection matrix can be fed into our algorithm and then a more discriminative solution for the projection matrix is obtained, resulting in higher classification performance.

The main difference with existing work on discriminant data representation is the joint use of three key points. The first is that the projection matrix contains two different types of discriminative features, namely inter-class sparsity and robust LDA. Second, the optimization of the proposed global criterion initializes the projection matrix with a hybrid solution. And finally, a gradient descent approach is used in the optimization scheme to tune this hybrid solution for the projection matrix.

The experiments conducted show that the proposed method resulted in an improvement in classification accuracy in the majority of tested cases and was able to outperform several competing methods. The rest of the paper is organized as follows. Section 2 describes related work and presents the notations used. Section 3 presents the criterion and solution details of the proposed method along with two initialization procedures. The obtained experimental results are presented in Sect. 4. Finally, Sect. 5 concludes the paper.

## 2 Related work and notations

In this section, we describe some related works and briefly introduce the gradient descent method and how we used it to obtain a better embedding space by selecting the best and most relevant features of the data. In addition, we will show how the introduction of the $\ell_{2,1}$ [38] norm and inter-class sparsity constraint was used for feature selection and helped in discrimination [26], and enumerate some recent methods that have used such a constraint by embedding it in their global criterion to ensure that the method performs feature selection [9, 18].

## 2.1 Notations

We will start by introducing the notations that we use in our paper. We will refer for the training set by $\mathbf{X} = [\mathbf{x}_1, \mathbf{x}_2, ..., \mathbf{x}_N] \in \mathbb{R}^{d \times N}$, with $d$ the dimension of each sample.

Each sample $\mathbf{x}_j$ is a column vector with $d$ features $\in \mathbb{R}^d$.

The number of training samples will be denoted by $N$. In addition, $C$ will represent the total number of classes. The Frobenius norm of a matrix $\mathbf{Z} \in \mathbb{R}^{d \times N}$ is obtained through the following formula $\|\mathbf{Z}\|_F = \sqrt{\sum_{k=1}^{d} \sum_{j=1}^{N} z_{ij}^2}$. The $\ell_{2,1}$ norm of a matrix $\mathbf{Z} \in \mathbb{R}^{d \times N}$ is obtained through the following formula $\|\mathbf{Z}\|_{2,1} = \sum_{k=1}^{d} \sqrt{\sum_{j=1}^{N} z_{ij}^2}$, and the $\ell_2$ norm for the vector $\mathbf{z} = [z_1, z_2, ..., z_d]$ is obtained as follows: $\|\mathbf{z}\|_2 = \sqrt{\sum_{k=1}^{d} z_i^2}$.

Table 1 shows the main notations used in our paper.

## 2.2 Related work

Recently, many feature extraction methods have been proposed. Some of these methods have built-in constraints that implement feature ranking/selection in the method and rank the features of their projection matrices. Feature selection or ranking is becoming more and more a trending problem in machine learning. Very often, using all data features does not lead to high classification performance. Feature selection aims to efficiently select the most relevant features of the data that best describe the data and improve discrimination [25, 32, 39, 40]. On the other hand, feature extraction aims to derive new sets of features from the original ones. The derived features are usually more discriminative than the original ones.

The best known method to tackle the curse of high dimensionality is the principal component analysis (PCA) [24] method. PCA is an unsupervised feature extraction method that transforms the features of the original data and projects them into a low-dimensional space. In the well-known supervised linear discriminant analysis (LDA) [8, 27] method, label information is required for the data. LDA and its variants are among the most widely used and discriminative algorithms in machine learning. LDA estimates a projection matrix in which the desired space maximizes the variance between classes and minimizes the variance within classes. The projection axis $\mathbf{w}$ would be the solution to the Fisher criterion [14]:

$$\mathbf{w} = \arg \min_{\mathbf{w}^T \mathbf{w} = 1} \mathbf{w}^T (\mathbf{S}_w - \lambda \mathbf{S}_b) \mathbf{w} \tag{1}$$

where $\lambda$ is a small positive constant that balances the effect of the two scatter matrices (within-class scatter matrix $\mathbf{S}_w$ and between-class scatter matrix $\mathbf{S}_b$), which could be calculated as:

$$\mathbf{S}_b = \frac{1}{N} \sum_{i=1}^{C} n_i (\mu_i - \mu) (\mu_i - \mu)^T \tag{2}$$

$$\mathbf{S}_w = \frac{1}{N} \sum_{i=1}^{C} \sum_{j=1}^{n_i} (\mathbf{x}_j^i - \mu_i) (\mathbf{x}_j^i - \mu_i)^T \tag{3}$$

where $\mu$, $\mu_i$ are the mean of all data samples and the mean of samples of the $i$th class, respectively. Many variants of LDA were proposed and still being proposed (e.g. [5, 45, 46]), as the linear discriminant analysis showed good interpretability for the data.

### 2.2.1 Review of robust sparse linear discriminant analysis (RSLDA):

RSLDA [35] was proposed to address many limitations of classical LDA [27], RSLDA mainly adds $\ell_{2,1}$ regularization to the projection matrix. This regularization term is added to the global criterion to ensure that the method performs feature ranking and weighting. RSLDA minimizes the following criterion:

$$\min_{\mathbf{P,Q,E}} Tr \left( \mathbf{Q}^T \mathbf{S} \mathbf{Q} \right) + \lambda_1 \|\mathbf{Q}\|_{2,1} + \lambda_2 \|\mathbf{E}\|_1 \tag{4}$$

$$s.t. \quad \mathbf{X} = \mathbf{P} \mathbf{Q}^T \mathbf{X} + \mathbf{E}, \ \mathbf{P}^T \mathbf{P} = \mathbf{I}$$

where $\mathbf{Q} \in \mathbb{R}^{d \times d}$ and $\mathbf{P} \in \mathbb{R}^{d \times d}$ denote the projection matrix and the orthogonal matrix, respectively. $\mathbf{E} \in \mathbb{R}^{d \times N}$ is the error matrix. $\mathbf{S} \in \mathbb{R}^{d \times d}$ is the difference matrix $(\mathbf{S}_w - \lambda \mathbf{S}_b)$, and $\lambda_1$ and $\lambda_2$ are two regularization parameters that balance the importance of the different terms. In the criterion of RSLDA, the $\ell_{2,1}$ norm was imposed on the projection matrix to achieve feature selection.

**Table 1** Main notations used in the paper

| Notation | Description | |
|---|---|---|
| $\mathbf{X}$ | Training data samples $\in \mathbb{R}^{d \times N}$ | |
| $\mathbf{X}_i$ | Training data samples in the $i$th class $\in \mathbb{R}^{d \times n_i}$ | |
| $\mathbf{P}$ | Orthogonal matrix $\in \mathbb{R}^{d \times d}$ | |
| $\mathbf{Q}$ | Projection matrix $\in \mathbb{R}^{d \times d}$ | |
| $\mathbf{D}$ | Diagonal matrix | $\in \mathbb{R}^{d \times d}$ |
| $\mathbf{S}_w$ | Within-class scatter matrix | $\in \mathbb{R}^{d \times d}$ |
| $\mathbf{S}_b$ | Between-class scatter matrix | $\in \mathbb{R}^{d \times d}$ |
| $d$ | Dimensionality of data | |
| $N$ | Number of training samples | |
| $n_i$ | Number of samples in the $i$th class | |
| $C$ | Number of classes | |
| $\mathbf{x}_i$ | The $i$th data sample $\in \mathbb{R}^d$ | |

### 2.2.2 Review of inter-class sparsity least square regression:

In [37], the authors propose the inter-class sparsity-based discriminative least square regression ICS_DLSR [37]. This method provides a linear mapping to the soft label space, where the dimension of the latent space is set to the number of classes. This method was able to construct a model in which the margins of samples from the same class are greatly reduced, while the margins for samples from different classes are increased. This was achieved by adding a class-wise row-sparsity constraint for the transformed features. ICS_DLSR minimizes the following problem:

$$\min_{\mathbf{Q}, \mathbf{E}} \frac{1}{2} ||\mathbf{Y} + \mathbf{E} - \mathbf{Q}\mathbf{X}||_F^2 \\ + \frac{\lambda_1}{2} ||\mathbf{Q}||_F^2 + \lambda_2 \sum_{i=1}^{C} ||\mathbf{Q}\mathbf{X}_i||_{2,1} + \lambda_3 ||\mathbf{E}||_{2,1} \tag{5}$$

where $\mathbf{X} \in \mathbb{R}^{d \times N}$ is the training set with $N$ samples from $C$ classes. $\mathbf{Y} \in \mathbb{R}^{C \times N}$ is a binary label matrix corresponding to the training set. $\mathbf{Q}$ is the projection matrix and $\mathbf{E}$ denotes the errors. $\lambda_1$, $\lambda_2$ and $\lambda_3$ are three regularization parameters.

Another similar method is the one described in [26], where the $\ell_{2,1}$-norm is applied to the transformation of the original linear discriminant analysis.

## 3 Proposed method

In this section, we present our problem formulation and show the steps applied to solve it. Our method is mainly considered as a linear projection method used for feature extraction, aiming at finding a more discriminative projection matrix. Two variants of the method are proposed. These two variants differ in the initialization step. Our proposed method has adopted feature ranking by using the solution of RSLDA as the initial estimate for the sought transformation. The next step is to fine tune the initial guess for the projection matrix by minimizing the proposed criterion with a gradient descent method, which aims to find the required solution of the projection matrix $\mathbf{Q}$.

The gradient descent algorithm is one of the simplest and most efficient algorithms for solving unconstrained optimization problems. In our algorithm, we have used the gradient descent approach to compute the projection matrix $\mathbf{Q}$ and find the solution.

### 3.1 Formulation

The main goal of our approach is to obtain both the projection matrix $\mathbf{Q} \in \mathbb{R}^{d \times d}$ and the orthogonal matrix $\mathbf{P} \in \mathbb{R}^{d \times d}$ using a unique criterion. In fact, the main contribution consists of the following objective function:

$$f(\mathbf{Q}, \mathbf{P}) = Tr\left(\mathbf{Q}^T \mathbf{S} \mathbf{Q}\right) + \lambda_1 \sum_{i=1}^{C} ||\mathbf{Q}^T \mathbf{X}_i||_{2,1} \\ + \lambda_2 ||\mathbf{X} - \mathbf{P} \mathbf{Q}^T \mathbf{X}||_F^2 \tag{6}$$

$$s.t. \quad \mathbf{P}^T \mathbf{P} = \mathbf{I}$$

where $\mathbf{X}_i \in \mathbb{R}^{d \times n_i}$ is the data matrix belonging to the $i$th class, $n_i$ is the number of training samples in the $i$th class, and $C$ is the number of classes.

The first term in equation (6) is the LDA criterion, where $\mathbf{S}$ is the LDA scatter matrix, which can be calculated as $\mathbf{S} = \mathbf{S}_w - \lambda \mathbf{S}_b$, where $\mathbf{S}_b$ is the between-class matrix and $\mathbf{S}_w$ is the within-class matrix. These two matrices are given by equations (2) and (3), respectively. This first term provides the first type of data discrimination in the projected space. The second term of the criterion is to ensure that transformed features of the same class share a common sparse structure in the projected space. $\mathbf{Q}$ is the projection matrix we are looking for. The second term provides the second type of data discriminant in the projected space. In addition, a variant of the (PCA) constraint is introduced to guarantee that the original data are well recovered. This third term improves the quality of the sample discrimination, as shown in [35]. $\lambda_1$ and $\lambda_2$ are two trade-off parameters that can be used to control the importance of the different terms. It is known that the $\ell_{2,1}$-norm of a matrix can be written as:

$$||\mathbf{Z}||_{2,1} = Tr\left(\mathbf{Z}^T \mathbf{D} \mathbf{Z}\right), \tag{7}$$

where $\mathbf{D}$ is a diagonal matrix that is given by:

$$\mathbf{D} = \begin{pmatrix} \dfrac{1}{||\mathbf{z}(1)||_2 + \epsilon} & \cdots & 0 \\ 0 & \ddots & 0 \\ 0 & 0 & \dfrac{1}{||\mathbf{z}(d)||_2 + \epsilon} \end{pmatrix}, \tag{8}$$

where $\mathbf{Z}(j)$ represents the $j$th row of $\mathbf{Z}$.

By substituting the second term of the criterion by its trace form shown in equation (7), problem (6) can be viewed as:

$$f(\mathbf{Q}, \mathbf{P}) = Tr\left(\mathbf{Q}^T \mathbf{S} \mathbf{Q}\right) + \lambda_1 \sum_{i=1}^{C} Tr\left(\left(\mathbf{Q}^T \mathbf{X}_i\right)^T \mathbf{D}_i \mathbf{Q}^T \mathbf{X}_i\right)$$
$$+ \lambda_2 \|\mathbf{X} - \mathbf{P} \mathbf{Q}^T \mathbf{X}\|_F^2$$

(9)

$$\min_{\mathbf{Q}, \mathbf{P}} f(\mathbf{Q}, \mathbf{P}) \ s.t. \quad \mathbf{P}^T \mathbf{P} = \mathbf{I}.$$

Equation (9) represents the criterion for the proposed method. The minimization of the first term of this criterion is targeting a projection matrix that ensures class discrimination with linear discriminant analysis (LDA). The second term of the criterion is introduced to obtain class sparsity. By introducing this condition, the transformed features from each class obtain a common sparse structure. Finally, a variant of the "principal component analysis" constraint is introduced in our proposed criterion [10]. This last constraint was introduced to maintain the energy-preserving property of PCA, and this constraint ensures the robustness of our data.

To find a solution for the proposed method, we used the gradient descent algorithm. Gradient descent algorithm is a mathematical process used for minimizing a particular function. The gradient algorithm uses the first derivative of the objective function given in Eq. 9. The gradient algorithm allows the person to solve the optimization problem in such a way that one knows the gradient from a particular point and can move in that direction to get a solution. The use of gradient algorithm has many advantages, and we mention the most important of them as follows:

- It has less computational complexity compared to other methods. Finding the solution by the descent gradient algorithm is usually less computationally expensive. Using the descent gradient to find a solution results in a faster model.
- It leads to accurate solutions. The gradient algorithm leads to a more accurate solution to the minimization problem than the closed-form solution.

### 3.2 Solution steps to the proposed method

To solve the problem formulated above, we adopted the alternating direction method of multipliers (ADMM) [1] and calculated each variable, while other variables are fixed as follows:

- **Calculate the orthogonal matrix P: P** can be calculated by fixing the variable **Q** and through solving the following equation:

$$\min_{\mathbf{P}^T \mathbf{P} = \mathbf{I}} \|\mathbf{X} - \mathbf{P} \mathbf{Q}^T \mathbf{X}\|_F^2.$$

(10)

Using $\mathbf{P}^T \mathbf{P} = \mathbf{I}$ and the fact that the squared norm of a matrix $\mathbf{A}$ is given by $\|\mathbf{A}\|_F^2 = Tr(\mathbf{A}^T \mathbf{A}) = Tr(\mathbf{A} \mathbf{A}^T)$, problem (10) is equivalent to the following maximization:

$$\min_{\mathbf{P}^T \mathbf{P} = \mathbf{I}} \|\mathbf{X} - \mathbf{P} \mathbf{Q}^T \mathbf{X}\|_F^2 \rightarrow \max_{\mathbf{P}^T \mathbf{P} = \mathbf{I}} Tr(\mathbf{P}^T \mathbf{X} \mathbf{X}^T \mathbf{Q}).$$

(11)

One can find a solution to equation (11) by performing singular value decomposition of $\mathbf{X} \mathbf{X}^T \mathbf{Q}$. Suppose the SVD decomposition is given by $SVD\left(\mathbf{X} \mathbf{X}^T \mathbf{Q}\right) = \mathbf{U} \Sigma \mathbf{V}^T$. Then, **P** is obtained as [46]:

$$\mathbf{P} = \mathbf{U} \mathbf{V}^T.$$

(12)

- **Calculate the Projection matrix Q:** Gradient descent is an iterative optimization technique used to minimize a function by moving in the direction of steepest descent in each iteration. The way the gradient method is used differs in different areas. In machine learning and classification, gradient is used to iteratively update the parameters of the desired model. We adopted the gradient descent method to compute **Q** in each iteration of the proposed method as follows: The orthogonal matrix **P** is fixed. Let us consider the trace form of the criterion of our problem:

$$f(\mathbf{Q}, \mathbf{P}) = Tr\left(\mathbf{Q}^T \mathbf{S} \mathbf{Q}\right) + \lambda_1 \sum_{i=1}^{C} Tr\left(\mathbf{X}_i^T \mathbf{Q} \mathbf{D}_i \mathbf{Q}^T \mathbf{X}_i\right)$$
$$+ \lambda_2 \|\mathbf{X} - \mathbf{P} \mathbf{Q}^T \mathbf{X}\|_F^2.$$

We calculate the gradient of the objective function w.r.t **Q** as follows:

$$\mathbf{G} = \frac{\delta f}{\delta \mathbf{Q}} = 2 \mathbf{S} \mathbf{Q} + \lambda_1 \sum_{i=1}^{C} 2 \mathbf{X}_i \mathbf{X}_i^T \mathbf{Q} \mathbf{D}_i$$
$$+ 2\lambda_2 \left[\mathbf{X} \mathbf{X}^T \mathbf{Q} - \mathbf{X} \mathbf{X}^T \mathbf{P}\right].$$

(13)

Using the gradient matrix, we can update **Q** by:

$$\mathbf{Q}_{t+1} = \mathbf{Q}_t - \alpha \mathbf{G}$$

(14)

where $\mathbf{Q}_{t+1}$ and $\mathbf{Q}_t$ denote the projection matrix **Q** in iteration $t + 1$ and iteration $t$, respectively. The step length (learning rate) is given by $\alpha$.

- **Update $\mathbf{D}_i$:** We update $\mathbf{D}_i$, $(i = 1, ..., C)$ by:

$$\mathbf{D}_i = \begin{pmatrix} \dfrac{1}{\left\|\mathbf{Q}^T \mathbf{X}_i(1)\right\|_2 + \epsilon} & \cdots & 0 \\ 0 & \ddots & 0 \\ 0 & 0 & \dfrac{1}{\left\|\mathbf{Q}^T \mathbf{X}_i(d)\right\|_2 + \epsilon} \end{pmatrix},$$

(15)

**Fig. 1** Graphical illustration of the proposed data representation framework. The transformations provided by RSLDA and ICS-DLSR are combined into a hybrid projection matrix $\mathbf{Q}_{Hybrid}$, which serves as input to our proposed algorithm. After minimizing the global criterion given in Eq. (9), we obtain the optimal projection matrices $\mathbf{Q}$ and $\mathbf{P}$. The final data representation is obtained using the projection matrix $\mathbf{Q}$

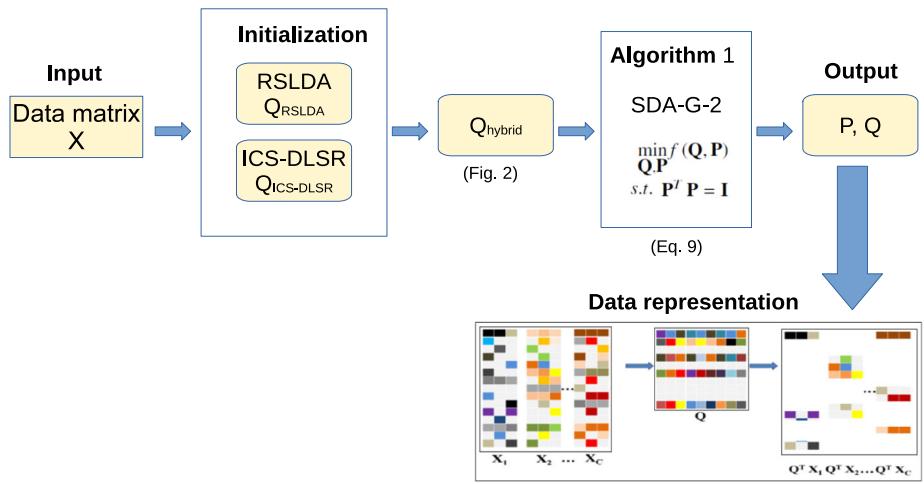

where $\epsilon$ is a small positive scalar and $\mathbf{Q}^T\mathbf{X}_i(j)$ represents the $j$th row vector of matrix $\mathbf{Q}^T\mathbf{X}_i$.

**Algorithm 1** summarizes our proposed method and describes the main steps for solving problem (6).

| Algorithm. 1. | Supervised discriminant analysis using gradient (SDA_G_1) |
|---|---|
| | Supervised discriminant analysis using gradient via combined initialization (SDA_G_2) |
| **Input:** | 1. Data samples $\mathbf{X} \in \mathbb{R}^{d \times N}$ |
| | 2. Labels of the training samples |
| | 3. The step length of the gradient descent $\alpha$ |
| | 4. Parameters $\lambda_1$, $\lambda_2$ |
| **Output:** | $\mathbf{P}$, $\mathbf{Q}$ |
| **Initialization:** | $\mathbf{Q}^{(0)}$ obtained from RSLDA or using a hybrid combination (see sect. 3.3). |
| **Process:** | set $t = 0$ and $\mathbf{Q} = \mathbf{Q}^{(0)}$ |
| | **Repeat** |
| | Fix $\mathbf{Q}$, update $\mathbf{P}^{(t+1)}$ using Eq. (12). |
| | Calculate the gradient matrix $\mathbf{G}$ using Eq. (13) |
| | Fix $\mathbf{P}$, update $\mathbf{Q}^{(t+1)}$ using Eq. (14). |
| | Update $\mathbf{D}_i$ using Eq. (15) |
| | set $t = t + 1$ |
| | **Until** convergence |

The projection of the training and test samples is carried out using the estimated projection matrix $\mathbf{Q}$. This is given by $\mathbf{z}_{train} = \mathbf{Q}^T \mathbf{x}_{train}$ and $\mathbf{z}_{test} = \mathbf{Q}^T \mathbf{x}_{test}$, where $\mathbf{x}_{train}$ is a training data sample and $\mathbf{x}_{test}$ is a test data sample.

## 3.3 Initialization of projection matrix Q

The linear transformation $\mathbf{Q}$ needs a good initial estimate, since it is estimated by a gradient descent update rule. In this section, we present two initialization procedures that lead to two variants of the proposed algorithm.

### 3.3.1 Using RSLDA algorithm

In this variant, the initial estimate $\mathbf{Q}^{(0)}$ for the linear projection matrix $\mathbf{Q}$ is given by the solution of the RSLDA [35] method (solved by a separate ADMM optimization). RSLDA was able to provide implicit feature selection by imposing the $\ell_{2,1}$ norm over the sought projection matrix. Moreover, the introduction of the error matrix helped in tracking and modeling the random noise. These introduced terms have resulted in RSLDA obtaining a discriminative and efficient transformation. The solution of our proposed method is computed using the gradient approach, which requires a good initial estimate to ensure good performance. By adopting the solution derived from RSLDA method, our proposed variant could adopt the advantages of this method. Figure (1) describes the initialization process using the projection matrix provided by RSLDA.

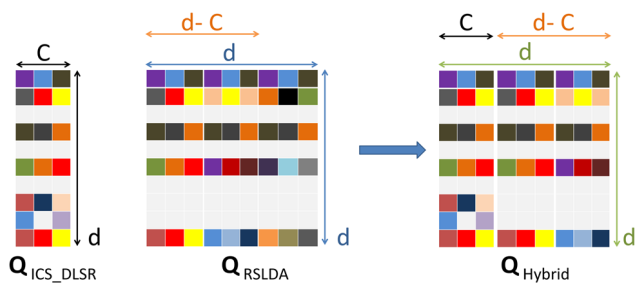

**Fig. 2** Combined initialization using the linear embeddings derived from ICS_DLSR and RSLDA

 

### 3.3.2 Hybrid combination of projection matrices obtained from the two embedding methods RSLDA and ICS_DLSR

In the second variant of our proposed algorithm, the initial projection matrix $\mathbf{Q}^{(0)}$ is set to a hybrid combination of the transformation matrices obtained by the two embedding methods RSLDA [35] and ICS_DLSR [37].

Let the number of rows of the hybrid transform $\mathbf{Q}^{(0)}$ be $d$. The number of columns (projection axes), on the other hand, can be set to any arbitrary value. Without loss of generality, to be consistent with linear methods, we will assume that the total number of columns of $\mathbf{Q}^{(0)}$ is $d$. Thus, $\mathbf{Q}^{(0)} \in \mathbb{R}^{d \times d}$. According to [37], the linear transformation $\mathbf{Q}_{ICS\_DLSR}$ obtained by the ICS_DLSR algorithm is $\in \mathbb{R}^{d \times C}$, where $d$ and $C$ represent the dimension of the features and the number of classes, respectively. On the other hand, the RSLDA method [35] has its own linear transformation $\mathbf{Q}_{RSLDA} \in \mathbb{R}^{d \times d}$. The sought initial hybrid projection matrix $\mathbf{Q}^{(0)}$ used in our algorithm is denoted by $\mathbf{Q}_{Hybrid}$. It is constructed by taking all $C$ columns of $\mathbf{Q}_{ICS\_DLSR}$ to which the first $d - C$ columns of $\mathbf{Q}_{RSLDA}$ are attached. The resulting projection matrix $\mathbf{Q}_{Hybrid}$ is $\in \mathbb{R}^{d \times d}$. The strategy for the hybrid initialization methodology is shown in Fig. 2.

In the above construction of the hybrid matrix $\mathbf{Q}_{Hybrid}$, our work fixed the number of projection axes for each projection type to $C$ and $d - C$ for ICS_DLSR and RSLDA, respectively. We emphasize the fact that these dimensions can be changed.

In our experiments, according to Table 2, we can see that the value of $C$ that represents the number of classes varies between 10 and 50 for the datasets used. The number of features for each dataset, $d$, is also shown in the same table.

### 3.4 Computational complexity

In this section, the computational complexity of the proposed method is analyzed (see **Algorithm 1**). From [35], we know that the computational complexity of the RSLDA method is $\mathcal{O}(\tau(d^2N + 4d^3))$, where $\tau$ denotes the number of iterations of RSLDA, $N$ denotes the number of samples, and $d$ denotes their dimension.

**Computational cost of the first variant.** Matrices $\mathbf{Q}$, $\mathbf{P}$, are sought to be calculated. The orthogonal matrix $\mathbf{P}$ requires a singular value decomposition. The computational cost of a decomposition of a $d \times N$ matrix would be $\mathcal{O}(N^3)$. $\mathbf{Q}$ is computed in the second step of the procedure, it requires the computation of the corresponding gradient matrix, but since these two steps consist only of simple matrix operations, they have low computational cost and therefore can be ignored. Also, the step provided for updating $\mathbf{D}_i$ from equation (15) is a simple matrix operation that has very low cost. Thus, the cost of one iteration of Algorithm 1 is $\mathcal{O}(N^3)$. Assuming that $\tau'$ represents the number of iterations of the proposed iterative scheme (Algorithm 1), the cost of Algorithm 1 is $\mathcal{O}(\tau'(N^3))$.

On the other hand, in the first variant of our proposed method, we used the RSLDA method for initializing the projection matrix $\mathbf{Q}$ before feeding it to our algorithm. Therefore, the complexity of the RSLDA method should be added to the complexity of our proposed method.

In summary, the total cost of the first variant would be the sum of RSLDA cost added to the cost of our proposed method, which equals $\mathcal{O}(\tau(d^2N + 4d^3)) + \mathcal{O}(\tau'(N^3))$.

**Computational cost of the second variant.** For the second proposed variant, we constructed the initial estimate of the projection matrix by combining solutions obtained using the RSLDA [35] and ICS_DLSR [37] methods. Since we know that the ICS_DLSR algorithm has a complexity of $\mathcal{O}(\tau(d^3))$, the total cost of the second proposed variant would be $\mathcal{O}(\tau(d^3)) + \mathcal{O}(\tau(d^2N + 4d^3)) + \mathcal{O}(\tau'(N^3))$.

**Table 2** Brief datasets description

| Dataset | Type | Number of Samples | Number of features | Number of classes | Descriptor |
|---|---|---|---|---|---|
| USPS | Digits | 1100 | 256 | 10 | RAW-brightness images |
| Honda | Face | 2277 | 1024 | 22 | RAW-brightness images |
| COIL20 | Object | 1440 | 177 | 20 | Local Binary Patterns |
| Extended Yale B | Face | 2414 | 1024 | 38 | RAW-brightness images |
| FEI | Face | 700 | 1024 | 50 | RAW-brightness images |
| MNIST | Digits | 60,000 | 2048 | 10 | Deep features (ResNet-50) |
| 20 News | Text | 2,000 | 100 | 4 | Term frequency times inverse document frequency |
| Tetra synthetic | Points | 400 | 100 | 4 | Coordinates |

# 4 Performance study

To test both variants of our proposed method, we conducted experiments on several datasets including faces, objects, and handwritten datasets. Detailed information on these datasets is presented in this section. Next, we are going to present the experimental setup and the results obtained.

## 4.1 Datasets

In our work, we have conducted our experiments over the following five public datasets in addition to a large-scale dataset: **USPS**[1] digits dataset, **Honda**[2] dataset, **COIL20**[3] object dataset, **Extended Yale B**[4] face dataset, **FEI**[5] dataset, and the large-scale **MNIST** dataset consisting of 60,000 images.

1. **USPS Digits Dataset**[6]: The US Postal Service or abbreviated as (USPS) [23] is a well-known handwritten digits dataset used for digit recognition. This dataset represents 10 digits (from 0 to 9), and it contains a total of 110 images for each digit, thus a total number of 1100 images in which the dimension of each one is 256. Raw brightness images are used for classification. Popular training percentages for this dataset are used as we use 30, 40, 55, and 65 image samples from each class as training samples and set the rest as test samples.

2. **Honda dataset**[7]: Honda dataset contains a total of 2277 face images that represent the faces of 22 different individuals in different conditions. Each class contains approximately 97 images. Popular training percentages are used as we use 10, 20, 30, and 50 image samples from each class as training samples and set the rest as test samples. Raw brightness images are used for the classification process.

3. **COIL20 object dataset**[8]: Columbia Object Image Library (COIL20) [19] dataset used for evaluation in our experiments consists of a total of 1440 images representing 20 different classes, and each class has 72 images. Different images of this dataset represent various objects in which each object is rotated around a vertical axis. As a training set, we used 20, 25, 30, and 35 image samples from each class and set the rest for testing. The image descriptor used is the local binary patterns (LBP) [17]. We used the uniform LBP histogram (59 values). Three LBP descriptors are constructed from the image using eight points and three values for the radius ($R$=1, 2, and 3 pixels). Thus, the final concatenated descriptor has 177 values.

4. **Extended Yale B face dataset**[9]: This dataset is a popular dataset used for image classification purposes [11]. The Extended Yale dataset is constructed from facial taken in different illuminations and facial expressions for each subject. The dataset we have used is the cropped version of the original Extended Yale B dataset, it contains between 58 and 64 images per class, and each class contains images that represent one individual. The total number of classes in this dataset is 38, and the total number of image samples is 2414. An adequate percentage of the training data is adopted as we have used 10, 15, 20, and 25 samples from each class for training, and the remaining were used as test samples. Each image of this dataset is rescaled to 32×32 pixels and represented through grayscale representation. Raw brightness images of dimension 1024 are used in the experiments.

5. **FEI dataset**[10]: The FEI face dataset contains 700 images of 50 students and staff members of FEI (14 images for each person). It is a face dataset that contains a set of colorful face images (images are resized to 32 × 32 pixels) taken against a white background. The images are in an upright frontal position with profile rotation of up to about 180 degrees. Raw brightness images of dimension 1024 are used. We used 5, 6, 7, and 8 image samples from each class as training samples. We should emphasize that the choice for these training set sizes comes from the fact that the number of samples in each class of the FEI dataset is relatively low (only 14) compared with other datasets.

6. **MNIST dataset:** The large-scale MNIST digits dataset is challenging. It contains a total number of 60,000 images representing 10 different classes. The length of the used image descriptor is 2048. The descriptor is obtained from the (ResNet-50)[11] convolutional neural network.

7. **20 News text dataset**: This is a cropped version of the 20 newsgroups dataset, with binary occurrence data for 100 words across 16,242 postings. This dataset

**Fig. 3** Some images of datasets

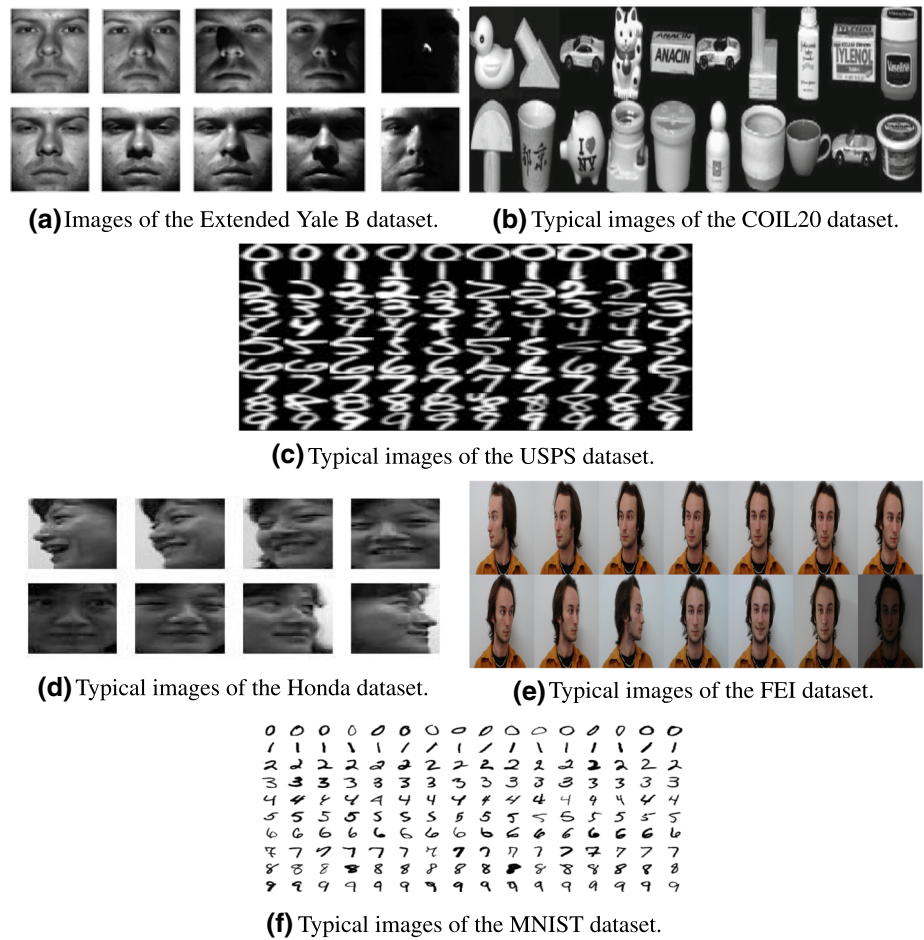

**(a)** Images of the Extended Yale B dataset.

**(b)** Typical images of the COIL20 dataset.

**(c)** Typical images of the USPS dataset.

**(d)** Typical images of the Honda dataset.

**(e)** Typical images of the FEI dataset.

**(f)** Typical images of the MNIST dataset.

contains a total of 2000 samples belonging to four classes.

8. **Tetra synthetic dataset:** The Tetra dataset was defined in [29, 30]. This dataset consists of 400 data points belonging to four classes. The data points are in $\mathbb{R}^3$, and this dataset presents the challenge associated with low inter-cluster distances.

Table 2 presents a summary for all the information concerning the datasets used in our paper.

## 4.2 Results

As already reported, the proposed method has two variants, namely:

- Supervised discriminant analysis using gradient technique (**SDA_G_1**): In this variant, our proposed method is implemented in the case that the initial projection matrix $\mathbf{Q}^{(0)}$ is set to the output of the RSLDA [35] algorithm as presented in Sect. 3.3.1.
- Supervised discriminant analysis using gradient via combined initialization (**SDA_G_2**): The second

variant of the proposed method consists of initializing the projection matrix $\mathbf{Q}^{(0)}$ as a hybrid combination of the solutions derived from the RSLDA [35] and ICS_DLSR [37] methods. The initial transformation construction is shown in Fig. 2 and detailed in Sect. 3.3.2.

The proposed variants have been compared with the following methods: K-nearest neighbors (KNN) [12], support vector machines (SVM) [3] (the linear SVM was implemented using the LIBSVM library)[12] linear discriminant analysis (LDA) [27], local discriminant embedding (LDE) [4], PCE [20] (unsupervised method) ICS_DLSR [37] and robust sparse LDA (RSLDA) [35].

All experiments for all compared methods were conducted under the same conditions to guarantee a fair comparison. For each compared embedding method, the whole dataset is randomly split into a training part and a test part.

First, for each compared method, a projection matrix is estimated from the training part. Then, training and test data are projected onto the new space using the already computed transformation. Finally, the classification of the test data is then performed using the nearest-neighbor classifier (NN) [6].

---

[12] https://www.csie.ntu.edu.tw/cjlin/libsvm/.

**Table 3** Mean classification performance (%) of the competing methods on the tested datasets

| Dataset\method | Training Samples | KNN | SVM | LDA | LDE | PCE | ICS_DLSR | RSLDA | SDA_G_1 | SDA_G_2 |
|---|---|---|---|---|---|---|---|---|---|---|
| USPS | 30 | 87.01 | 88.21 | 84.91 | 83.54 | 72.01 | 88.46 | 89.45 | 89.50 | **90.29** |
| | 40 | 88.56 | 90.40 | 86.19 | 85.3 | 72.30 | 90.16 | 91.11 | **91.81** | 91.46 |
| | 55 | 90.51 | 92.09 | 88.64 | 87.16 | 73.32 | 91.25 | 92.65 | **93.07** | 92.87 |
| | 65 | 91.76 | 93.16 | 89.29 | 88.58 | 74.11 | 91.53 | 92.89 | **93.71** | 93.49 |
| Honda | 10 | 64.12 | 71.32 | 65.95 | 65.74 | 61.86 | 70.79 | 69.90 | 70.16 | **72.14** |
| | 20 | 77.69 | 83.60 | 79.39 | 79.25 | 75.33 | 82.95 | 83.03 | 83.60 | **84.64** |
| | 30 | 84.78 | 89.09 | 85.84 | 86.24 | 82.55 | 88.20 | 89.04 | 89.41 | **90.12** |
| | 50 | 91.36 | 94.15 | 92.28 | 92.34 | 90.03 | 93.53 | 94.13 | 94.53 | **95.10** |
| FEI | 5 | 88.98 | 91.18 | 92.60 | 90.67 | 86.04 | 92.16 | 93.19 | 93.81 | **94.58** |
| | 6 | 90.35 | 92.93 | 94.18 | 92.15 | 88.73 | 93.65 | 94.25 | 94.75 | **95.08** |
| | 7 | 92.60 | 94.31 | 95.60 | 94.26 | 91.09 | 95.20 | 95.66 | 96.20 | **96.29** |
| | 8 | 94.27 | 95.23 | 96.03 | 95.57 | 93.20 | 96.17 | 96.43 | **96.97** | 96.40 |
| COIL20 | 20 | 94.58 | 97.65 | 96.19 | 95.00 | 94.87 | **98.04** | 96.73 | 96.89 | 97.66 |
| | 25 | 95.79 | 98.22 | 97.07 | 96.12 | 95.99 | 98.22 | 97.74 | 97.89 | **98.59** |
| | 30 | 96.65 | 98.70 | 97.81 | 97.01 | 97.49 | 98.75 | 98.26 | 98.52 | **99.08** |
| | 35 | 97.14 | 98.81 | 98.15 | 97.42 | 98.11 | 99.12 | 98.68 | 98.80 | **99.39** |

Different sizes of training sets were used. Moreover, for a given percentage of training data, the whole evaluation is repeated 10 times. That means that we adopt 10 random splits for every configuration and report the average recognition rate (rate of correct classification of test part) over these 10 random splits.

We used PCA as a preprocessing technique. In our experiments, PCA [24] is used as a dimensionality reduction technique and used to preserve 100% of the data's energy. Concerning the parameter α, we should set it to a small value. In our experiments, this value was chosen in $\{10^{-7}, 10^{-5}\}$.

The obtained results are summarized in Table 3. This table depicts the classification performance of the proposed variants in addition to the competing methods using the USPS, Honda, FEI, and COIL20 datasets. The results are obtained using different training and testing percentages from the data. The results shown in this table are obtained using the nearest-neighbor classifier. Table 4 presents the obtained classification performance using the Extended Yale B dataset. In this table, various training percentages corresponding to a different number of samples used in the training process are shown. We should emphasize that more competing methods are presented in table 4, and these additional methods are ELDE, in addition to SULDA and MPDA. These methods were added to enrich the comparison using more methods. To further improve the comparison over the Extended Yale B dataset, we have added more methods to the comparison table, based on low-rank representations. The added methods are the low-rank linear regression (LRLR) [2], low-rank ridge regression (LRRR) [2], sparse low-rank regression (SLRR) [2], and the low-rank preserving projection via graph regularized reconstruction (LRPP GRR) [36]. Low-rank-based methods findings can be found in the bottom part of table 4. The depicted rates are the average over 10 random splits and correspond to different numbers of training samples. The first column inside the table depicts the number of training images per class.

Table 5 illustrates the classification performance for the competing methods alongside the proposed variants using the large-scaled MNIST dataset that contains a total number of 60,000 images in total. Results shown in this table are obtained using one split while using 1000 samples from each class for training and the remaining samples were used for testing.

Table 6 depicts the obtained classification performance using the News20 text dataset. The results presented in this table are the mean classification obtained using 10 split while using 20% and 30% of the data samples from each class for training and the remaining samples were used for testing.

Figure 5 presents the obtained recognition rate (%) associated with the LDA [27], LDE [4], and RSLDA [35] in addition to the two proposed variants of our method. The recognition rate is plotted as a function of the dimension of the projected features. The results are shown for (a) the COIL20 dataset, (b) the Extended Yale B, and (c) the HONDA dataset. 30, 10, and 10 samples from each class are used for training, respectively. The depicted results were obtained using the **nearest-neighbor** (NN) classifier.

**Table 4** Mean classification performance (%) of using the Extended Yale B dataset

| DatasetMethod | Training Samples | KNN | SVM | LDA | LDE | ELDE | PCE | SULDA | MPDA | ICS_DLSR | RSLDA | SDA_G_1 | SDA_G_2 |
|---|---|---|---|---|---|---|---|---|---|---|---|---|---|
| Ext. Yale B | 10 | 69.80 | 73.85 | 82.32 | 79.92 | 85.85 | 86.39 | 84.61 | 83.67 | 86.56 | 86.79 | 87.10 | **88.42** |
| | 15 | 75.20 | 80.02 | 86.76 | 83.77 | 89.30 | 89.23 | 88.72 | 86.82 | 89.53 | 89.93 | 90.04 | **91.21** |
| | 20 | 80.24 | 85.79 | 90.7 | 88.44 | 93.07 | 92.19 | 91.66 | 90.38 | 93.14 | 93.59 | 93.75 | **93.81** |
| | 25 | 82.24 | 89.03 | 92.17 | 90.43 | 94.09 | 93.35 | 92.14 | 91.79 | 94.50 | 94.92 | 95.02 | 95.09 |
| | | LRLR | SLRR | LRPP_GRR | LRRR | | | | | | | | |
| | 10 | 84.63 | 87.95 | 84.82 | 87.76 | | | | | | | | |
| | 15 | 86.31 | 89.75 | 89.07 | 91.09 | | | | | | | | |
| | 20 | 88.93 | 92.58 | 91.42 | 93.19 | | | | | | | | |
| | 25 | 90.98 | 94.24 | 92.25 | **95.51** | | | | | | | | |

We have used the results obtained from 21 evaluations and using six different datasets from the experiments conducted in this paper to study the statistical analysis of our proposed method. We performed the Friedman test [7] and computed the critical distance CD. The obtained results of the conducted test yield to the conclusion that the tested methods do not have the same performance. Figure 4 shows the CD diagram for the nine methods including our two proposed variants, where the average rank of each is marked along the axis.

**Experiments using synthetic data:**

In addition to the image datasets, we also conducted some experiments on the synthetic Tetra dataset [28]. This dataset consists of 400 data points belonging to four classes. The original data points of this dataset are in $\mathbb{R}^3$, but in our experiments, the dimension was augmented to 100 so each data sample is represented by 100 features. The three-dimensional dataset is transformed to a high-dimensional dataset $\in \mathbb{R}^{100}$ using a random projection matrix.

This dataset was chosen because it presents the challenge associated with low inter-cluster distances. The distance between the clusters is minimal. Data points of Tetra are visualized in Fig. 6. One can see that the clusters nearly touch each other.

Figure 7 illustrates the TSNE visualization of the projected samples of the Tetra dataset using the original linear discriminant analysis (LDA) and RSLDA in addition to the first variant of our suggested method **SDA_G_1**. By looking at that figure, it is noticeable that our method provides very good class separation properties and leads to the most compact representation among competing methods. The proposed method ensured superior performance when it is applied on datasets with low inter-cluster distances.

### 4.3 Analysis of parameter sensitivity

In this section, we investigate the effect of changing the proposed method's parameters on the classification performance using different datasets. The proposed method has mainly two parameters to be configured, namely $\lambda_1$ and $\lambda_2$.

Figure 8 shows the variation of the classification performance when adopting different parameter combinations of the proposed method using the Extended Yale B and Honda datasets. Figure 8a and 8c shows the variation of the classification rates using the Extended Yale B and Honda datasets in the cases of using 10 and 20 training samples from each class, respectively, using the first variant of the proposed method **SDA_G_1**. Also, the classification performance is studied on the same datasets with adopting the same training percentages using the second variant of the proposed method **SDA_G_2**. The corresponding results are depicted in Fig. 8b and d.

**Table 5** Mean classification accuracies (%) of different methods on the tested datasets

| Dataset \Method | Training Samples | KNN | SVM | LDA | LDE | PCE | ICS_DLSR | RSLDA | SDA_G_1 | **SDA_G_2** |
|---|---|---|---|---|---|---|---|---|---|---|
| MNIST | 1000 | 91.75 | 97.58 | 85.74 | 93.22 | 93.77 | 98.02 | 97.95 | 98.21 | **98.33** |

**Table 6** Classification Performance (%) on the News20 text dataset

| News20 | | | | |
|---|---|---|---|---|
| | Training Percentage | | | |
| | 20% | | 30% | |
| Method | Classification accuracy | Method | Classification accuracy |
| LDA | 68.04 | LDA | 68.70 |
| RSLDA | 68.11 | RSLDA | 68.88 |
| **SDA_G_1** | 68.38 | **SDA_G_1** | 69.10 |
| **SDA_G_2** | **68.87** | **SDA_G_2** | **69.58** |

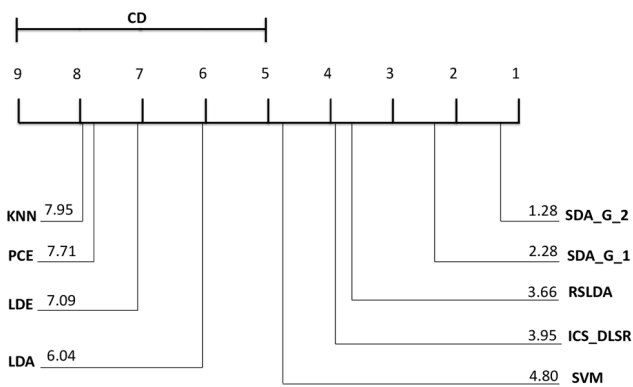

**Fig. 4** Statistical Analysis—CD diagram

For the Extended Yale B dataset, we monitored the classification performance obtained by both of our proposed variants using different values for $\lambda_1$ and $\lambda_2$. $\lambda_1$ and $\lambda_2$ were varied for the ranges from $[10^{-5}, 1]$ and $[10^{-3}, 10]$ respectively. We noticed that satisfactory rates for the Extended Yale B dataset can be obtained when $\lambda_1$ was chosen from the range $[10^{-3}, 10^{-1}]$ and $\lambda_2$ within the range of $[10^{-2}, 10^{-1}]$.

Similar to the Extended Yale B experiment, we studied the classification performance of the proposed schemes over the Honda dataset. We varied $\lambda_1$ in the range of $[10^{-3}, 10^3]$ and $\lambda_2$ in the range $[10^{-4}, 10^3]$. We noticed that satisfactory rates using Honda dataset can be obtained by choosing the value $\lambda_1$ from the range of $[10^{-1}, 10]$ and $\lambda_2$ from the range of $[10^{-3}, 10^2]$. We concluded that the values of the parameters $\lambda_1$ and $\lambda_2$ should lie in the intervals shown in the figures above to obtain satisfactory results using the proposed method. A value of 0.1 for both $\lambda_1$ and $\lambda_2$ seems to be a good choice for the two variants and the two datasets.

Figure 8 shows the variation of the classification performance according to the change of the parameters $\lambda_1$ and $\lambda_2$. This figure corresponds to the variants of the proposed method when applied on the Extended Yale B and Honda dataset using 10 and 20 samples from each class for training respectively and the rest for testing.

### 4.4 Analysis of results

From our analysis of the experiments conducted, we can make the following observations:

1. The classification performance obtained by the proposed method alongside the competing methods demonstrates that our proposed approach has outperformed competing methods in the majority of the cases.

2. The first proposed variant **SDA_G_1** has slightly outperformed the RSLDA method. This seems to be very realistic since the first proposed method mainly provides a fine-tuning of the RSLDA transformation.

3. In general, the second proposed scheme **SDA_G_2** is superior to the first proposed one **SDA_G_1**. This is explained by the fact that the second scheme benefits from the hybrid combination of two different powerful embedding methods as well as from the refinement provided by the gradient descent tool.

**Fig. 5** Classification accuracy (%) vs. dimension for different datasets

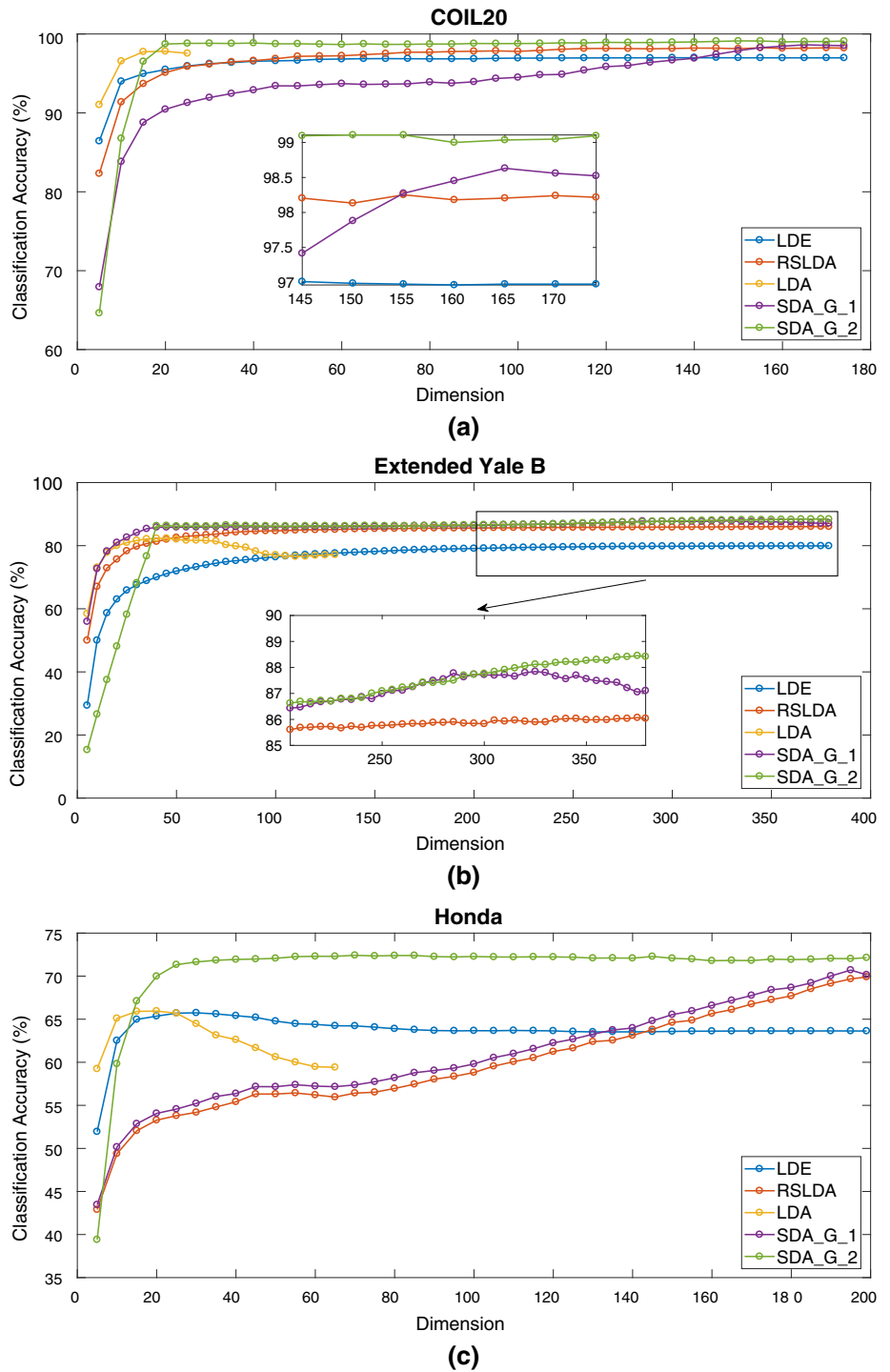

**(a)**

**(b)**

**(c)**

4. The proposed method proved superior performance using several types of image datasets, including faces, objects, and digits. Also, our approach demonstrated superior performance using a text dataset.

5. The proposed method showed superiority and led to very good class separation properties when it is applied on datasets with low inter-cluster distances.

6. The optimal parameters of the proposed method, which gives the best classification performance, have large ranges. In other words, the best classification performance is guaranteed most of the time by searching a small number of parameter combinations.

7. The competing method ICS_DLSR has performed better than our proposed method in a particular case using the COIL20 dataset while using 20 images from each class as training samples. On the other hand, the proposed method generally outperformed it using all other training percentages for the same dataset.

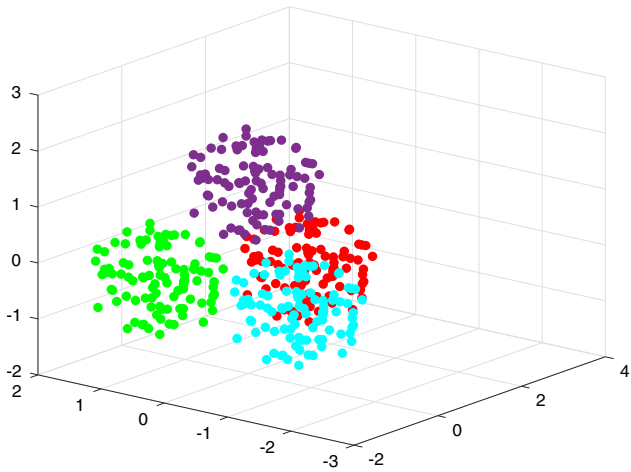

**Fig. 6** Visualization of the Tetra dataset points in the original space. These 3D points belong to four large full spheres close to each other

8. When the hybrid initialization was used in our algorithm, we adopted a combination of the two best-tuned transformation matrices obtained from the two methods RSLDA and ICS_DLSR as the initial transformation. In the majority of the tested cases, this has led to a noticeable enhancement in classification performance. The two best-tuned transformation matrices refer to the transformation matrices computed by two methods using the best parameter combination, which leads to the optimal performance of the method. It is worth noting that the use of the combination of the two tuned transformation matrices is not necessarily the best option for a combination in our framework. Other combinations may lead to better discrimination.

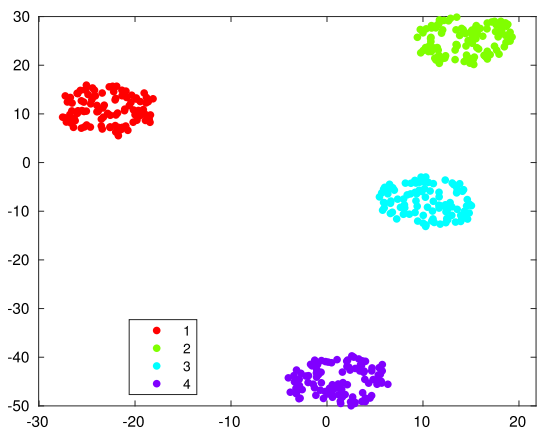

**(a)** Visualization of the projected samples of the Tetra dataset using Original LDA.

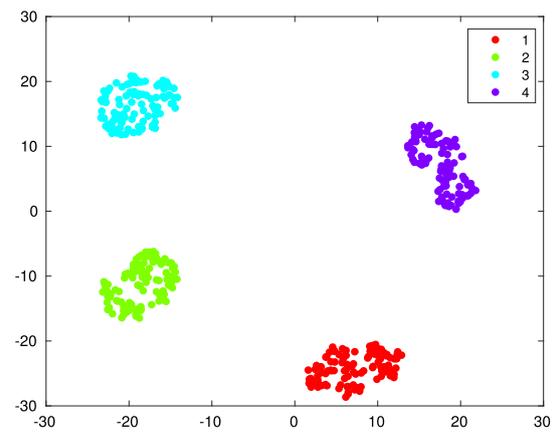

**(b)** Visualization of the projected samples of the Tetra dataset using RSLDA.

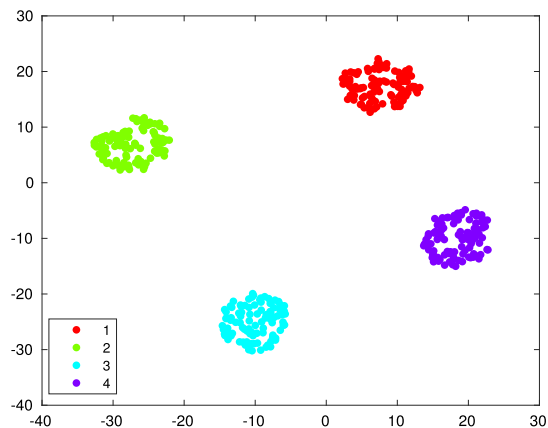

**(c)** Visualization of the projected samples of the Tetra dataset using **SDA_G_1**.

**Fig. 7** TSNE visualization of the projected samples of the Tetra dataset using LDA, RSLDA, and the first proposed variant **SDA_G_1**

**Fig. 8** Classification accuracy (%) according to parameters

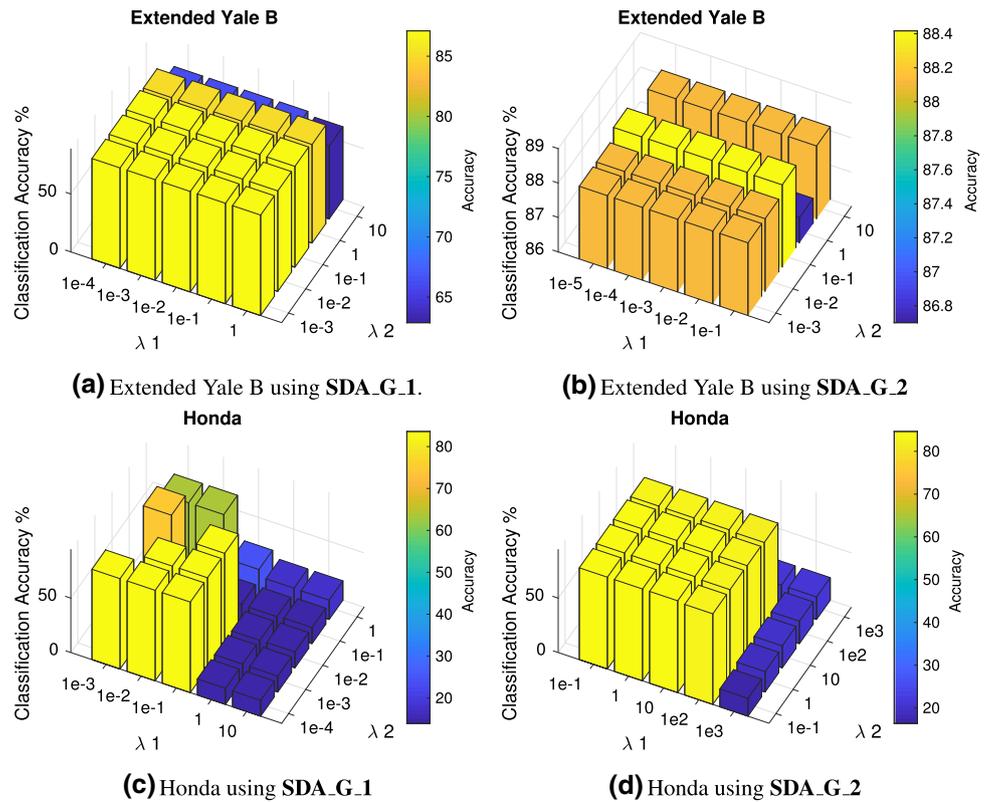

(a) Extended Yale B using **SDA_G_1**.

(b) Extended Yale B using **SDA_G_2**

(c) Honda using **SDA_G_1**

(d) Honda using **SDA_G_2**

Thus, the obtained classification performance using the second variant of our suggested approach (Table 3) could be further enhanced if other combinations for the initialization are used.

# 5 Conclusion

In this work, we introduced a novel criterion to obtain a discriminant linear transformation. This transformation efficiently integrates two different mechanisms of discrimination which are the inter-class sparsity and robust discriminant analysis. We deployed an iterative alternating minimization scheme to estimate the linear transformation and the orthogonal matrix associated with the robust LDA. In the alternating optimization, the linear transformation is efficiently updated via the steepest descent gradient technique.

We proposed two initialization variants for the linear transformation. The first scheme sets the initial solution to the linear transformation obtained by robust sparse LDA method (RSLDA). The second variant initializes the solution to a hybrid combination of the two transformations obtained by RSLDA and ICS_DLSR methods. The two variants of the proposed method have demonstrated superiority over competing methods and led to a more discriminative projection matrix, hence better classification performance. The main difference with existing work on discriminant data representation is the joint use of three key points. The first is that the projection matrix contains two different types of discriminative features, namely inter-class sparsity in addition to robust LDA. Second, the optimization of the proposed global criterion initializes the projection matrix with a hybrid solution. And finally, a gradient descent approach is used in the optimization scheme to tune this hybrid solution for the projection matrix.

The proposed framework is generic in the sense it allows the combination and tuning of other linear discriminant embedding methods.

Deep learning can provide a powerful data representation or classifier. However, this paper is about building a shallow discriminant model that can be trained with a few examples, e.g., a few dozen images. On the other hand, a deep learning model may need a large number of examples to provide a good data representation. It is worth mentioning that the presented projection method can also work with deep features, in the sense that the descriptors of the images can be provided by a deep model and the projection model can be provided by the proposed model. In other words, the proposed model can be used as a projection head of a pre-trained deep neural network.

**Funding** Open Access funding provided thanks to the CRUE-CSIC agreement with Springer Nature.

## Declarations

**Conflict of Interest Statement** The authors declare that the research was conducted in the absence of any commercial or financial relationships that could be construed as a potential conflict of interest.